\documentclass{article}
\usepackage[preprint]{corl_2026} 
\usepackage{amsmath}    
\usepackage{amssymb}    
\usepackage{graphicx}
\usepackage{algorithm}
\usepackage{algpseudocode}
\usepackage{caption}
\usepackage{float}
\usepackage[T1]{fontenc}
\graphicspath{{../Fig/}}
\title{CIDER: Continual Interactive Distillation for Embodied Reinforcement Learning}
\author{%
\textbf{Houlin Li }\textsuperscript{1,2, $*$} \quad
\textbf{Minghui Xu}\textsuperscript{1,2, $*$} \quad
\textbf{Guo Xu}\textsuperscript{1, \textdagger} \quad
\textbf{Xuan Du}\textsuperscript{1} \\
\textbf{Xiaohan Yan}\textsuperscript{1} \quad
\textbf{Chun Wang}\textsuperscript{1} \quad
\textbf{Yuxiang Yan}\textsuperscript{1} \quad
\textbf{Shukai Yang}\textsuperscript{1} \\ 
\textbf{Yongcheng Liu}\textsuperscript{1,\textdagger} \quad
\textbf{Wei Shan}\textsuperscript{1,\textdagger} \quad
\textbf{Maoqing Yao}\textsuperscript{1,\textdagger}
\\
\textsuperscript{1}AgiBot \quad
\textsuperscript{2}Shanghai Jiao Tong University  \quad
\\
\textsuperscript{$*$}These authors contributed equally to this work. \quad
\textsuperscript{\textdagger}Corresponding Author
}

\begin{document}
\maketitle
\begin{abstract}
    Human-in-the-loop real-world reinforcement learning enables rapid acquisition of effective robotic manipulation policies for individual tasks, often within tens of minutes. Yet it remains unclear how to extend this paradigm to continual learning, where a single policy must acquire new skills without losing previously learned behaviors. Existing real-world continual learning methods do not explicitly constrain prior behaviors, leading to severe catastrophic forgetting. We introduce Continual Interactive Distillation for Embodied Reinforcement Learning (CIDER), a continual reinforcement learning framework that freezes the accumulated historical policy as a teacher before learning each new task and interleaves task learning with distillation-based retention. We further introduce gradient routing to separate the gradients used for acquiring new tasks from those used for preserving prior behaviors. We evaluate our method with a single shared actor on six real-world household and industrial manipulation tasks. Interactive Distillation maintains high measured success on previously learned tasks across our six-task real-robot sequence while acquiring each new task in 10 to 20 minutes, whereas every baseline forgets at least one previous task. Additional ablations reveal the key design choices that govern the tradeoff between stability and plasticity in real-world continual reinforcement learning.
\end{abstract}
\section{Introduction}
Real-world reinforcement learning (RL) has become an effective mechanism for acquiring policies for high-performance robotic manipulation tasks \cite{luo2025precise, zhao2025realworldreinforcementlearningsuboptimal}, but it is unclear how such single-task skills should be accumulated over time. As objects, goals, and physical scenarios change, a deployable system must acquire new skills while continuing to execute previously learned ones \cite{traore2019continual,traore2019discorl,wolczyk2021continual}. This turns high-performing single-task RL into a different problem: not merely how to learn one expert policy, but how to learn a single deployable policy whose skills grow over time.

A straightforward approach is to continue fine-tuning the same policy with online RL as each new task arrives. Unfortunately, this procedure generally does not produce a policy that can perform all tasks seen so far \cite{wolczyk2021continual, 10.5555/3454287.3454319, wolczyk2022disentangling, huang2021continual, chen2024stable}. The actor and representation updates that improve the current task can overwrite the action distributions required by previous tasks, leading to catastrophic forgetting \cite{traore2019discorl, 10.5555/3454287.3454319, kaplanis2019policy, ahn2024reset}. This failure mode is especially costly in real-world RL, where recovering a previous skill may require additional robot interaction, model retraining, and repeated evaluation. Existing alternatives, including replay, regularization, and traditional distillation-based methods, mitigate parts of this problem \cite{traore2019discorl, wolczyk2021continual, 10.5555/3454287.3454319, ahn2024reset, yadav2024learning, zhang2023replay, kessler2022same, gai2023oer}, but continual real-world RL still needs a mechanism that constrains previous-task behavior throughout current-task HIL-RL without eliminating plasticity for the incoming task.

In this work, we propose Continual Interactive Distillation for Embodied Reinforcement Learning (CIDER), a training framework that interleaves online RL for the current task with expert distillation for previously learned tasks. When a new task arrives, the robot continues to collect real-world interaction and optimize the current policy with Human-in-the-loop (HIL) RL, preserving the plasticity and reward-driven improvement of RL training. In parallel, the frozen accumulated teacher that consolidates all previous tasks provides supervised action targets on stored state, constraining the current policy to preserve previous behaviors. After the current task is solved, the resulting policy is frozen as a new teacher. Unlike purely offline multi-expert consolidation, the same policy remains the acting learner throughout the task stream, and skill acquisition and retention occur within the same training process.
\begin{figure}[t]
    \centering
    \includegraphics[width=0.95\linewidth]{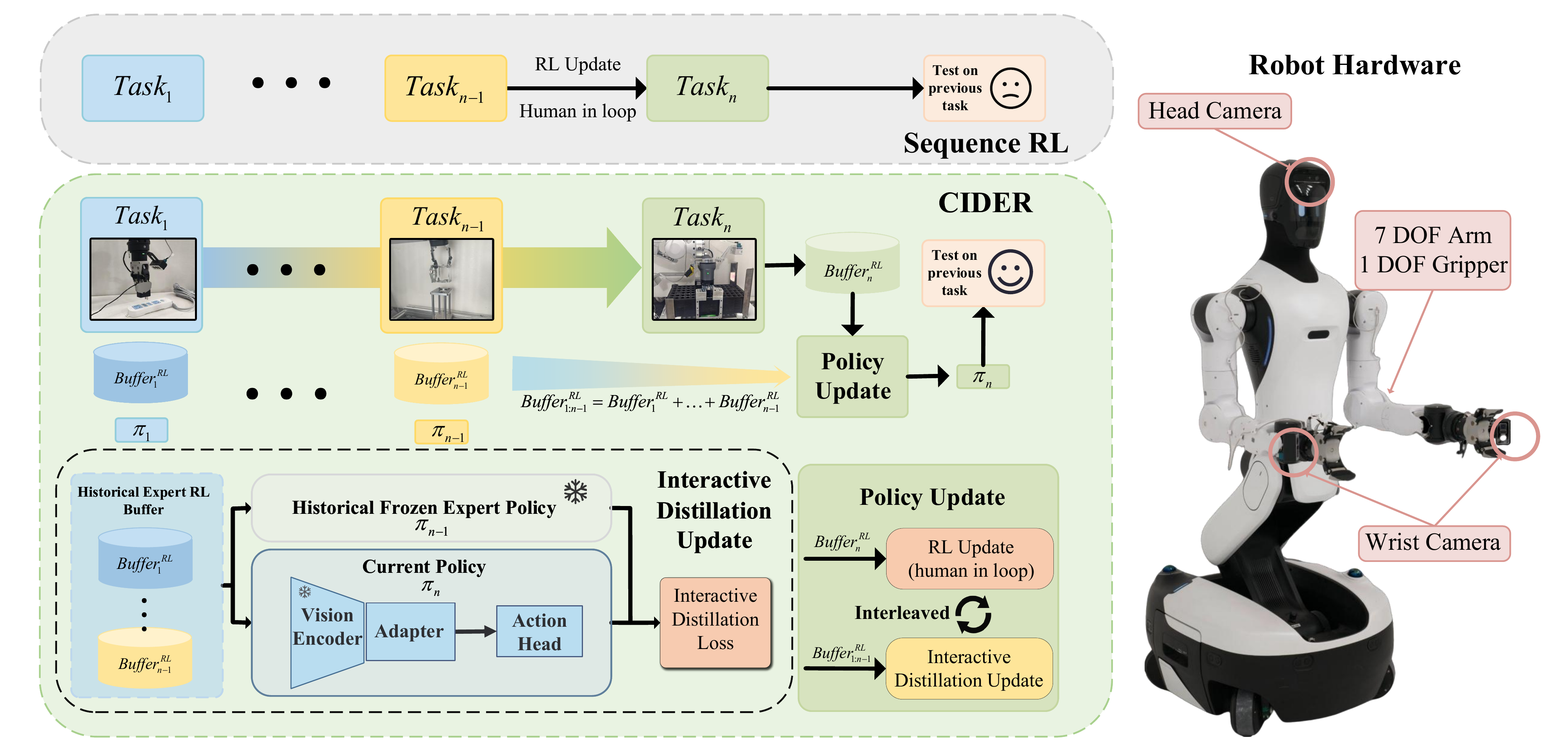}
    \vspace{-6pt}
    \caption{Overview of CIDER. Naive sequential RL forgets previous skills, while Interactive Distillation interleaves current-task online RL with rehearsal with a frozen expert on states from previous tasks.}
    \vspace{-18pt}
    \label{fig:overview}
\end{figure}

Our main contribution is to formalize CIDER as a practical framework for studying continual real-world robot RL in settings where individual manipulation tasks can already be solved reliably by human-in-the-loop RL. We instantiate CIDER in a visuomotor actor-critic system and evaluate it on a sequence of six short-horizon real-world manipulation tasks. Across this task suite, CIDER substantially reduces measured forgetting compared with sequential fine-tuning, replay, regularization, offline updates, and post-hoc expert consolidation baselines. These results suggest that online RL can serve as an acquisition mechanism for continual robot learning when paired with behavior-level retention from a frozen accumulated teacher, while leaving open questions about longer horizons, more ambiguous task conditioning, and larger task streams.
\section{Related Work}

\paragraph{Robotic HIL-RL.}
HIL-RL has emerged as a promising paradigm for complex robotic manipulation in the real-world, capable of achieving
near-human-level performance on highly dexterous tasks
\cite{luorlif,kumar2024applications,wu2023human}.
HIL algorithms introduce human intervention during policy execution,
allowing operators to provide timely corrections to robot actions;
these corrections provide the RL algorithm with high-quality experience
in the early stages of training, accelerating convergence and improving
final policy performance
\cite{luo2025precise,zhao2025realworldreinforcementlearningsuboptimal,luo2024serl,chen2025conrft}.
Our focus is not to develop a stronger single-task HIL-RL algorithm, but to study how such systems can be extended to continual learning, where new skills are acquired online while previously learned skills are retained without additional interaction on previous tasks.

\paragraph{Continual reinforcement learning and catastrophic forgetting.}
Continual RL (CRL) studies agents that continually adapt to new tasks rather than converging once to a fixed policy \cite{ring1994continual, khetarpal2022towards}. Unlike repeatedly applying standard RL across tasks, CRL requires balancing plasticity for acquiring new tasks and stability for retaining previous behaviors \cite{pan2026surveycontinualreinforcementlearning, zuffer2025advancementschallengescontinualreinforcement, arczewski2024reinforcement, NEURIPS2023_9d8cf124, 10.1561/2200000116, abbas2023loss}. Sequential updates can overwrite the knowledge needed by previous tasks, causing catastrophic forgetting \cite{chen2024stable, kaplanis2019policy, li2021sler, chen2022neural}. Existing methods mitigate this problem through soft regularization, e.g., EWC \cite{kirkpatrick2017overcoming}, SI \cite{zenke2017continual}, and MAS \cite{aljundi2018memory}, which constrain important parameters but do not directly preserve old action distributions; gradient projection, e.g., OGD \cite{farajtabar2020orthogonal} and GPM \cite{saha2021gradient}, which reduces interference but can shrink the update subspace as tasks accumulate \cite{lintrgp, zeng2019continual, deng2021flattening, yang2023data, wang2024training}; and structural isolation, e.g., PackNet \cite{mallya2018packnet}, Progress \& Compress \cite{schwarz2018progress}, and HAT \cite{serra2018overcoming}, which protects task-specific parameters but often requires masks, task identity, or growing capacity \cite{Rusu2016ProgressiveNN,wortsman2020supermasks, mallya2018piggyback}. These ideas have also been extended to robotic manipulation \cite{huang2021continual, kaplanis2019policy, yadav2024learning}. 

\paragraph{Policy distillation for skill consolidation.}
Policy distillation transfers the capabilities of expert policies to a student policy. In continual RL, previously task-specific experts can serve as teachers to enable ongoing consolidation of the model's capabilities. Existing work explores distillation for continual learning from several angles: training task-specific RL teachers and distilling their soft action distributions into a single student \cite{traore2019continual,traore2019discorl}, applying distillation to task-agnostic CRL, distributed teacher aggregation, or reset-and-distill pipelines that decouple a periodically reset online learner from a long-term offline actor \cite{Hafez2024,gu2026coevolvingpolicydistillation,ahn2024reset}. Policy distillation has been used both to accelerate the acquisition of new skills \cite{li2026continual} and to effectively retain the capabilities of previously learned policies \cite{li2021sler,zhao2024experience}. This work builds on these ideas and proposes Interactive Distillation, which interleaves online RL on the current task with frozen-teacher distillation on previous tasks, enabling continuous acquisition of new skills while preserving previously learned behaviors.
\section{Preliminaries}

\subsection{Sequential Real-World Continual RL}

We study continual RL for real-world visual manipulation, where a robot encounters a sequence of physical scenarios $\mathcal{M}_{1:T}=\{\mathcal{M}_1,\ldots,\mathcal{M}_T\}$. We write each scenario as $\mathcal{M}_i=(\mathcal{S}_i,\mathcal{A},P_i,R_i,\gamma)$, where $\mathcal{S}_i$ denotes the policy input space induced by camera states, object configuration, robot state, and optional task or goal specification; $\mathcal{A}$ is a continuous action space shared across scenarios; $P_i$ and $R_i$ are the scenario-specific transition kernel and reward; and $\gamma\in(0,1)$ is the discount factor. For visual policies, $\mathcal{S}_i$ should be understood as the observation space available to the policy rather than a fully observed Markov state.

At stage $k$, the robot interacts only with the current physical scenario $\mathcal{M}_k$. Once a previous scenario $\mathcal{M}_i$ has been completed, the learner cannot assume further access to its physical environment. It may only retain bounded information from previous scenarios, such as a transition archive $\mathcal{B}_i$. We denote the collection of transition archives available before stage $k$ by $\mathcal{B}_{<k}=\{\mathcal{B}_i\}_{i<k}$. After stage $k$, the goal is to produce a single deployable policy $\pi_\theta$ that performs well on all scenarios seen so far,
i.e. to maximize $\frac{1}{k}\sum_{i=1}^{k} J_i(\pi_\theta)$, where $J_i(\pi)=\mathbb{E}_{\tau\sim(\pi,P_i)}[\sum_{t=0}^{\infty}\gamma^t R_i(s_t,a_t)]$ and $\tau$ denotes a trajectory with states $s_t$ and actions $a_t$.

\subsection{Single Task Human-in-the-Loop Actor-Critic Learning}
\label{sec:prelim_sac}

For each active scenario $\mathcal{M}_k$, we use a HIL-RL pipeline with an off-policy actor-critic learner. The single-task learner maintains a stochastic actor $\pi_\theta(a|s)$, twin critics $Q_{\phi_1},Q_{\phi_2}$, and an active replay buffer $\mathcal{D}_k$ containing autonomous robot transitions, demonstrations, and human corrections. Human corrections are stored as additional off-policy experience, allowing the robot to recover from failures while continuing to improve through reward-driven interaction.

We instantiate the RL update with RLPD, which applies SAC-style actor-critic updates to the mixed replay buffer $\mathcal{D}_k$. For a transition $(s,a,r,s')\sim\mathcal{D}_k$, where $r$ is the sampled reward and $s'$ is the next state, we sample $a'\sim\pi_\theta(\cdot|s')$ and use the critic target $y=r+\gamma(\min_{q\in\{1,2\}}Q_{\bar{\phi}_q}(s',a')-\alpha\log\pi_\theta(a'|s'))$. The critic loss is $\mathcal{L}_{Q}(\phi_q)=\mathbb{E}_{(s,a,r,s')\sim\mathcal{D}_k}[(Q_{\phi_q}(s,a)-y)^2]$, and the actor is updated with $\mathcal{L}_{\mathrm{RL}}(\theta)=\mathbb{E}_{s\sim\mathcal{D}_k,\ a\sim\pi_\theta(\cdot|s)}[\alpha\log\pi_\theta(a|s)-\min_{q}Q_{\phi_q}(s,a)]$, where $\bar{\phi}_q$ denotes the target-network parameters and $\alpha$ controls the entropy-reward tradeoff.

In the continual setting, Bellman backups are applied only to the active replay buffer $\mathcal{D}_k$ from the current physical scenario. Historical transition archives are not used as transition data for value learning; only their state components are used for behavioral distillation.

\section{Interactive Distillation for Continual Real-World RL}

Interactive Distillation uses previously learned behavior as a constraint on the current policy during online learning. At stage $k$, the learner maintains the current buffer $\mathcal{D}_k$, historical transition archives $\mathcal{B}_{<k}$, and a frozen teacher policy $\pi_{\mathrm{T}}^{k-1}$ obtained after stage $k-1$. The actor continues to improve on the current scenario through online HIL-RL, while states extracted from archived transitions are used to query the behavior that should be retained from previous scenarios. The output of each stage is a single deployable policy $\pi_{\theta_k}$ and a new transition archive $\mathcal{B}_k$; the deployed policy remains one shared actor rather than a set of task-specific heads.

We use policy distillation to match the current student policy to the frozen teacher on states extracted from archived transitions. Let $\pi_{\mathrm{T}}(\cdot|s)$ denote a teacher action distribution and $\pi_\theta(\cdot|s)$ denote the current student. A generic distillation loss can be written as $\mathcal{L}_{\mathrm{distill}}(\theta)=\mathbb{E}_{(s,a,r,s')\sim\mathcal{B}}[\ell(\pi_\theta(\cdot|s),\pi_{\mathrm{T}}(\cdot|s))]$, where $\mathcal{B}$ denotes a generic transition archive and $\ell$ denotes the discrepancy between the student and teacher action distributions. For continuous control, both teacher and student are diagonal Gaussian policies, parameterized by $(\mu_{\mathrm{T}}(s),\sigma_{\mathrm{T}}(s))$ and $(\mu_\theta(s),\sigma_\theta(s))$, respectively.

\subsection{Stagewise Consolidation and Retention Objective}
\label{sec:stagewise_consolidation}

For $k=1$, no historical constraint is available, and the policy is trained with the HIL-RL procedure described in Section~\ref{sec:prelim_sac}. After the first scenario reaches its success criterion, we store a transition archive $\mathcal{B}_1$ and freeze the resulting actor as the first teacher $\pi_{\mathrm{T}}^1$.

For each subsequent scenario ($k>1$), training is initialized from the previously consolidated actor, $\pi_\theta \leftarrow \pi_{\theta_{k-1}}$, and the previous actor is frozen as the teacher, $\pi_{\mathrm{T}}^{k-1}=\operatorname{stopgrad}(\pi_{\theta_{k-1}})$. The current replay buffer is reset, and new transitions are collected only from the current scenario $\mathcal{M}_k$.

Training alternates between two update streams: HIL-RL updates on transitions from $\mathcal{D}_k$, followed by historical distillation updates on states extracted from $\mathcal{B}_{<k}$. We denote by $u_{\mathrm{RL}}$ and $u_{\mathrm{ID}}$ the number of updates assigned to the HIL-RL and distillation streams. At a high level, these updates loosely correspond to the objective
\[
\min_{\theta,\phi}
\mathcal{L}_{Q}(\phi;\mathcal{D}_k)
+ \mathcal{L}_{\mathrm{RL}}(\theta;\mathcal{D}_k)
+ \mathcal{L}_{\mathrm{distill}}^{k}(\theta),
\]
although the three terms are not evaluated on a common minibatch. The critic and the actor losses use transitions from the current scenario, whereas the distillation term uses the state components of archived transitions and teacher action distributions. In the main experiments, we set $u_{\mathrm{ID}}=u_{\mathrm{RL}}=1$ and include a sensitivity analysis over nearby ratios in Appendix~\ref{app:additional_ablations}.

\paragraph{Retention objective.}
Given the historical transition archives $\mathcal{B}_{<k}$ and the frozen teacher $\pi_{\mathrm{T}}^{k-1}$, we use reverse KL to match the current actor to the teacher on the state components of past transitions:
\[
\mathcal{L}_{\mathrm{distill}}^{k}(\theta)
=
\frac{1}{k-1}
\sum_{i=1}^{k-1}
\mathbb{E}_{(s,a,r,s')\sim\mathcal{B}_i}
\left[
D_{\mathrm{KL}}
\left(
\pi_{\theta}(\cdot|s)
\;\middle\|\;
\pi_{\mathrm{T}}^{k-1}(\cdot|s)
\right)
\right],
\]
where $D_{\mathrm{KL}}$ denotes the Kullback--Leibler divergence. The loss constrains the induced action distribution rather than the policy parameters themselves. This is useful for visuomotor policies, where parameter distance is only an indirect proxy for behavioral change on archived transitions.

\subsection{Optimization and Retention Details}
\label{sec:optimization_visual_actor_critic}

At the end of each scenario $i$, we construct a bounded transition archive
$\mathcal{B}_i=\mathrm{Archive}(\mathcal{D}_i)$. Before the training stage $k$, we evaluate the frozen teacher $\pi_{\mathrm{T}}^{k-1}$ once on the states extracted from the historical transition archives $\mathcal{B}_{<k}$ and cache its Gaussian action distributions: $\mathcal{C}_{<k}^{k-1}
=
\left\{
\left(s,\mu_{\mathrm{T}}(s),\sigma_{\mathrm{T}}(s)\right)
\mid (s,a,r,s')\in\bigcup_{i<k}\mathcal{B}_i
\right\}.$ Distillation updates then use the cached teacher outputs rather than repeatedly forwarding the teacher. Distillation minibatches are sampled in a task-balanced manner by drawing $i\sim\mathrm{Uniform}\{1,\ldots,k-1\}$ and then $(s,a,r,s')\sim\mathrm{Uniform}(\mathcal{B}_i)$; only the state component $s$ is used for the retention loss, so retention updates do not become dominated by any single previous scenario.

The critic is trained only on the current task's data $\mathcal{D}_k$, while archived transitions are excluded from Bellman backups. The actor receives gradients from both the current RL objective and the distillation retention objective:
\[
g_{\mathrm{RL}}
=
\nabla_\theta\mathcal{L}_{\mathrm{RL}}(\theta;\mathcal{D}_k),
\qquad
g_{\mathrm{distill}}
=
\nabla_\theta\mathcal{L}_{\mathrm{distill}}^{k}(\theta).
\]
%

Our visual policy includes a lightweight adapter shared by the actor and critic heads. By default, current-task actor gradients are blocked from the adapter, whereas historical distillation gradients update both the adapter and actor head. In our implementation, these gradients are applied with separate Adam optimizers. The complete Interactive Distillation procedure is summarized in Algorithm~\ref{alg:id} in Appendix~\ref{pseudocode}. The procedure interleaves online HIL-RL updates on the currently available scenario with historical distillation updates using cached teacher outputs.

Interactive Distillation preserves the loop of online adaptation of real-world RL but uses behavioral supervision from a frozen teacher instead of Bellman backups on old transitions. Compared with replay-based continual RL, the critic is trained only on transitions from the current scenario. Compared with post-hoc multi-expert distillation, consolidation occurs during the same stage in which the actor collects new real-world experience.
\section{Experiments}
We design the experiments to answer three questions. \textbf{Q1}. Can CIDER accumulate skills more reliably than alternative continual learning baselines?
\textbf{Q2}. Does sequential training provide positive transfer that improves learning efficiency on later tasks?
\textbf{Q3}. Which components of CIDER are most important for balancing plasticity on new tasks and stability on previous tasks?

\subsection{Real-Robot Task Suite and Continual Learning Protocol}
\begin{figure*}[t]
    \centering
    \includegraphics[width=\linewidth, trim=0 0 0 0, clip]{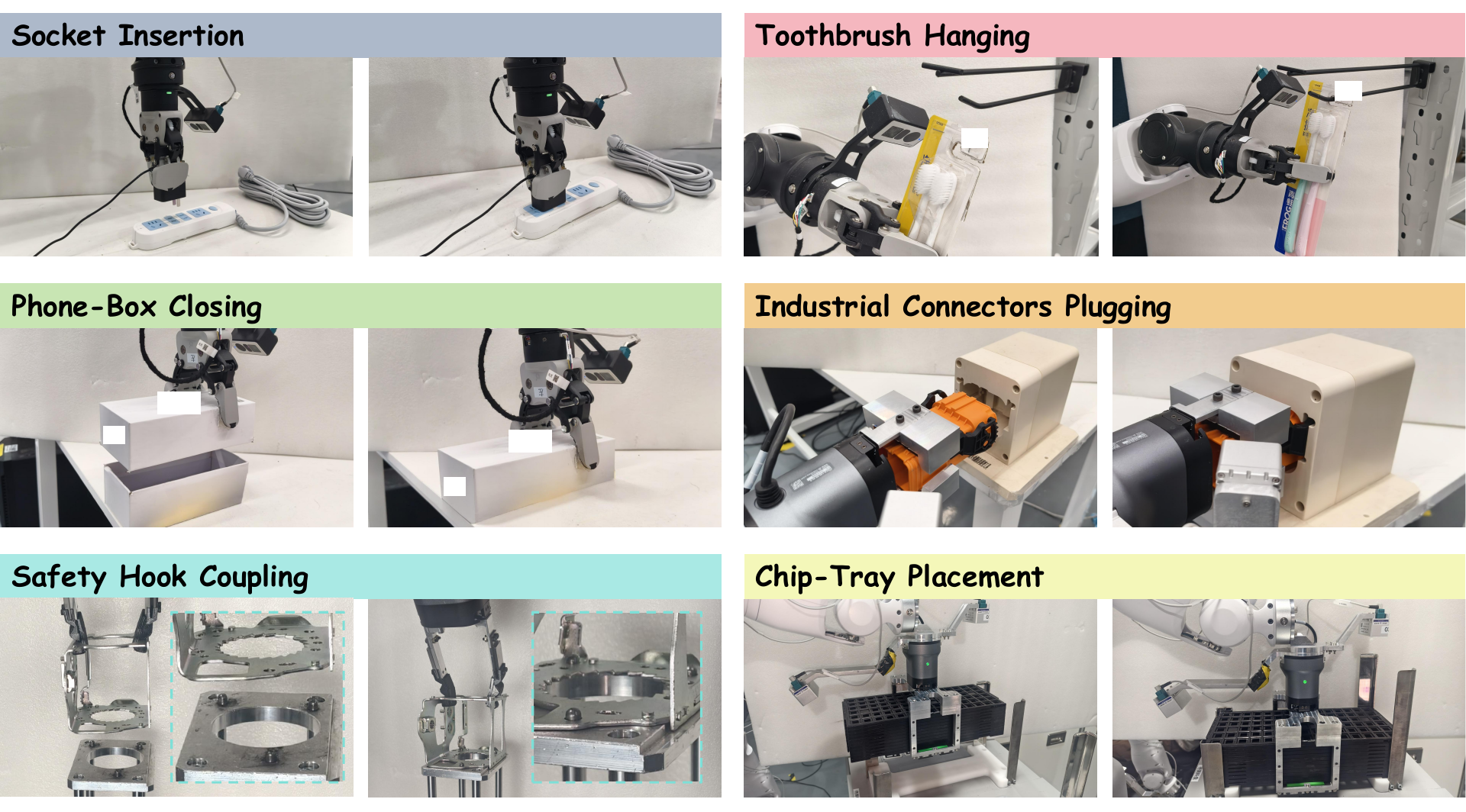}
    \vspace{-18pt}
    \caption{The six real-world manipulation tasks.}
    \vspace{-18pt}
    \label{fig:tasks}
\end{figure*}

\paragraph{Task suite and continual-learning protocol.}
We evaluate CIDER on a sequence of six real-world manipulation tasks spanning household and industrial scenarios, as shown in Figure~\ref{fig:tasks}. The household tasks include socket insertion, toothbrush hanging, and phone-box closing; the industrial tasks include industrial connectors plugging, safety hook coupling, and chip-tray placement. These tasks require visually guided contact-rich manipulation, precise alignment, and task-specific success conditions, while sharing the same robot platform, policy architecture, and continuous action space. This setting tests whether a single policy can accumulate diverse manipulation skills rather than specialize to a single scene.

The tasks are presented sequentially as $\mathcal{M}_1,\ldots,\mathcal{M}_T$, with $T=6$. During stage $k$, the learner can interact only with the current task $\mathcal{M}_k$. After each stage, the resulting policy $\pi_{\theta_k}$ is evaluated on all tasks seen so far, producing a lower-triangular continual-evaluation matrix:
$$
S_{i,k}=\mathrm{SuccessRate}(\pi_{\theta_k},\mathcal{M}_i),
\qquad 1\leq i\leq k\leq T.
$$

All success rates are measured over five independent sets of 20 real-robot trials per task. We report the average success over seen tasks after each stage $A_k=\frac{1}{k}\sum_{i=1}^{k}S_{i,k}$,
and the final average success $A_T$ after completing the full task sequence.

\paragraph{Single-task reference experts.}
Before evaluating continual learning, we train an independent expert for each task using the same human-in-the-loop real-world RL pipeline. We train one expert for each task to assess learnability, establish an upper reference for task performance, and disentangle catastrophic forgetting from intrinsic task difficulty. All six independently trained experts achieve 20 out of 20 successful trials, demonstrating that the task suite is solvable by the robot. This result indicates that failures in the continual learning setting arise from limitations of the training strategy rather than from infeasible task requirements.

\subsection{Main Results: Continual Skill Accumulation}
We compare CIDER against baselines that isolate different hypotheses about continual robot learning. \emph{Finetune} continues Human-in-the-Loop RL online on each new task from the previous policy parameters and discards the transitions of previous tasks. \emph{Finetune with Replay} additionally retains previous transitions in the replay buffer, testing whether replay alone is sufficient to prevent forgetting. \emph{Finetune with EWC} adds an elastic-weight-consolidation penalty to preserve weights estimated to be important for previous tasks. \emph{Sequential BC} replaces supervised fine-tuning with Behavior Cloning on the new task dataset. \emph{Sequential offline RL} replaces supervised fine-tuning with offline RL on the new task dataset. Finally, \emph{DisCoRL} trains independent experts for all tasks and then consolidates them into a single student in one offline distillation pass.

\begin{figure}[htbp]
    \centering
    \includegraphics[width=\linewidth]{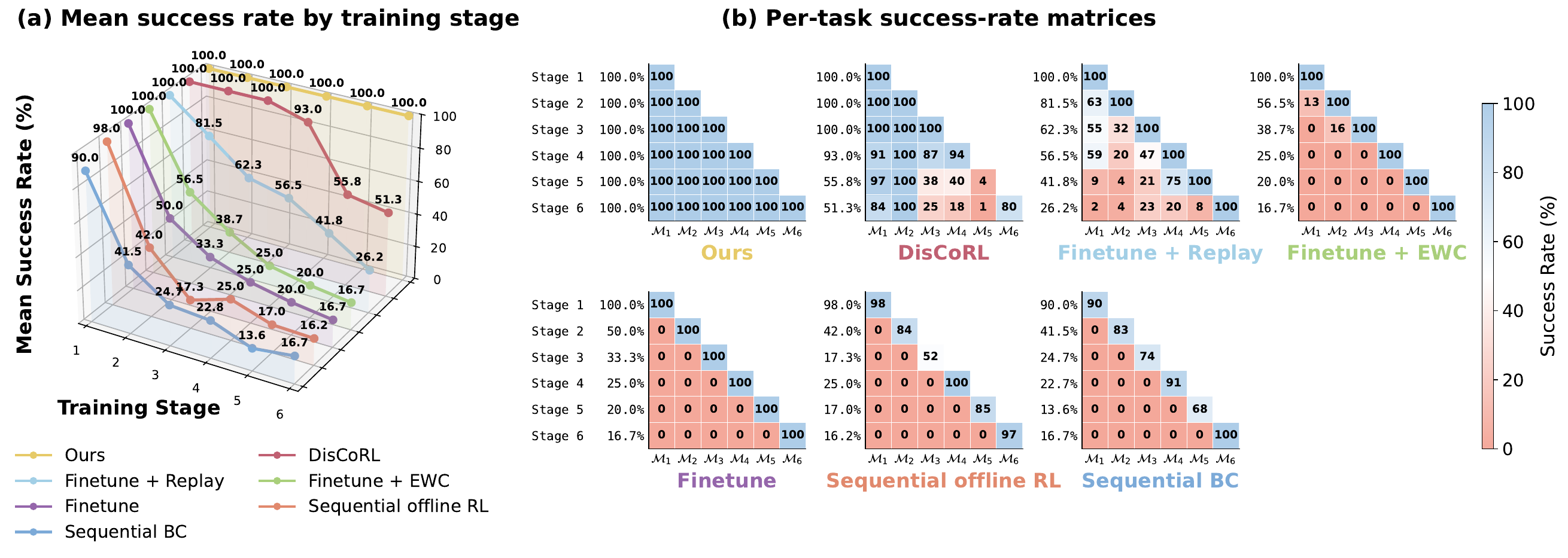}
    \vspace{-12pt}
    \caption{Continual-learning performance across the six-task sequence, evaluated over five independent evaluation sets with 20 rollouts each. \textbf{(a)} Mean success rate over all tasks seen so far after each training stage; \textbf{(b)} per-method success-rate matrices where row $k$ shows $S_{i,k}$ for each task $\mathcal{M}_i$ and the left margin reports the stage average $A_k$.}
    \vspace{-12pt}
    \label{fig:continual_results}
\end{figure}

The experiments in Figure~\ref{fig:continual_results} show that \emph{Finetune} learns the current task but catastrophically forgets previous skills after training on the next task. \emph{Finetune with Replay} retains old transitions in the replay buffer to slightly improve retention, but it still fails to preserve previous skills. \emph{Finetune with EWC} constrains the policy only through parameter importance rather than directly at the behavioral level. As a result, preserving important parameters alone is insufficient to prevent performance degradation. \emph{Sequential BC} and \emph{Sequential offline RL} also fail to maintain the previously learned tasks, suggesting that offline new-task updates or weight anchoring alone are insufficient for real-robot skill retention. \emph{DisCoRL} is a stronger baseline because it consolidates all independently trained experts at once. However, merging methods like \emph{DisCoRL} often ignore the interference between parameters of different models, resulting in performance drops when merging multiple models.

CIDER is the only evaluated method that preserves all six manipulation tasks across the sequential training process. As shown in Figure~\ref{fig:continual_results}, every policy obtained after stage $k$ succeeds on all tasks seen up to that point, yielding a final average success of 100\% over 100-trial evaluations.

\begin{figure}[htbp]
    \centering
    \includegraphics[width=0.85\linewidth]{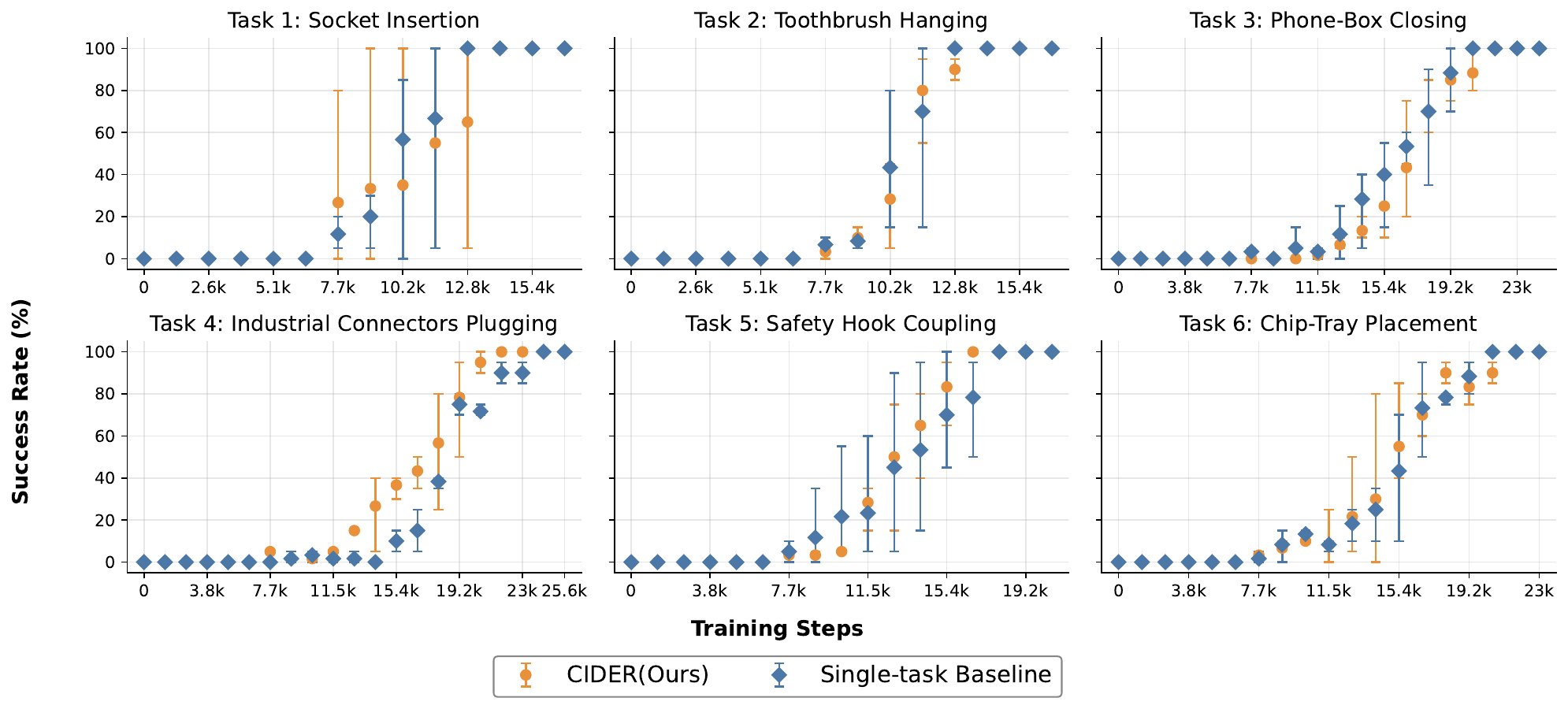}
    \vspace{-6pt}
    \caption{Success rate as a function of training steps for the CIDER policy (ours) and a single-task baseline across six manipulation tasks, averaged over three independent training runs per task.}
    \vspace{-15pt}
    \label{fig:forward_transfer} 
\end{figure}

\paragraph{Forward Transfer}
We further evaluate whether the accumulated policy provides forward transfer to new tasks by comparing the real-robot training steps required to reach the success criterion. For each evaluated task, CIDER starts from the previously consolidated policy, whereas the single-task baseline trains the same RL agent from the common initialization using only the current task; both settings use the same robot platform, reward, success criterion, and RL hyperparameters. As shown in Figure~\ref{fig:forward_transfer}, continuing from the consolidated policy requires no more training steps than learning each task independently from scratch across all evaluated tasks, reaches the success criterion faster on some tasks, and converges to successful behavior within approximately 10--20 minutes of real-robot training for each task. These results show that CIDER does not hinder new-task acquisition and modestly accelerates learning on some tasks, suggesting that the consolidated policy remains a useful warm start for subsequent real-robot learning despite variation in objects, contact dynamics, and success conditions.

\subsection{Ablations: Understanding CIDER}
\label{ablation}
We also perform ablations to identify which design choices are responsible for stable skill accumulation. First, we vary how gradients are routed through the visual adapter to determine whether the adapter should be updated by current-task RL, historical distillation, both, or neither. Second, we compare two student initialization schemes when learning the second task: \emph{Continue}, which initializes the student from the previously consolidated policy, and \emph{From scratch}, which reinitializes the student from a common parameterization shared across tasks while keeping the same current-task data and distillation setup. Additional ablations on the RL/distillation update ratio, historical-task sampling strategy, training task order and the loss type are deferred to Appendix~\ref{app:additional_ablations}.

\begin{figure}[htbp]
    \centering
    \includegraphics[width=\linewidth]{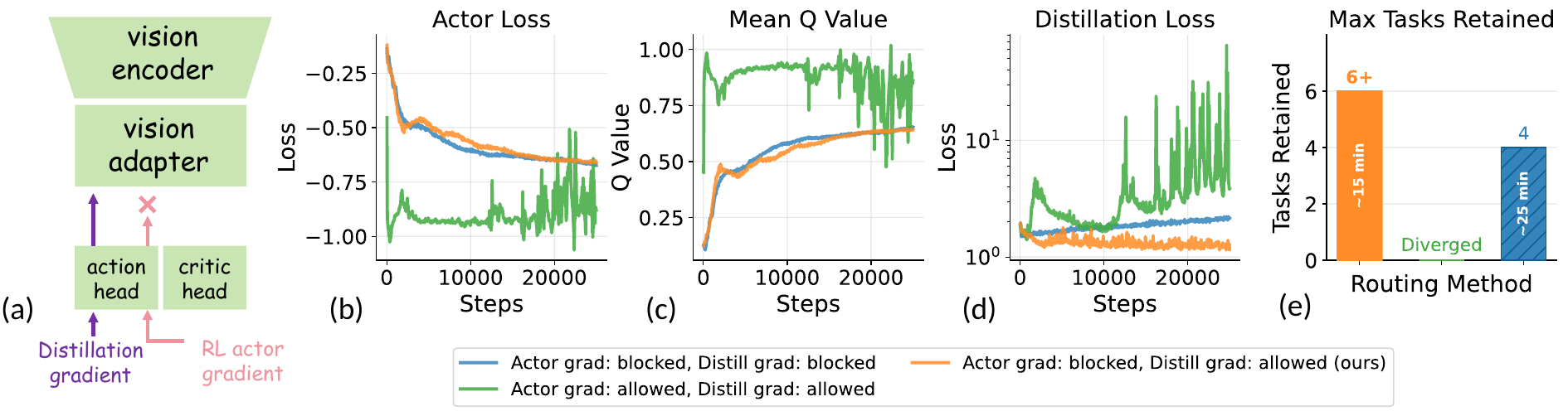}
    \vspace{-14pt}
    \caption{Adapter gradient routing ablation: actor loss, mean Q value, and distillation loss under three routing configurations.}
    \vspace{-18pt}
    \label{fig:adapter_routing}
\end{figure}

\begin{figure}[htbp]
    \centering
    \begin{minipage}[t]{0.64\linewidth}
        \vspace{0pt}
        \centering
        \includegraphics[width=\linewidth]{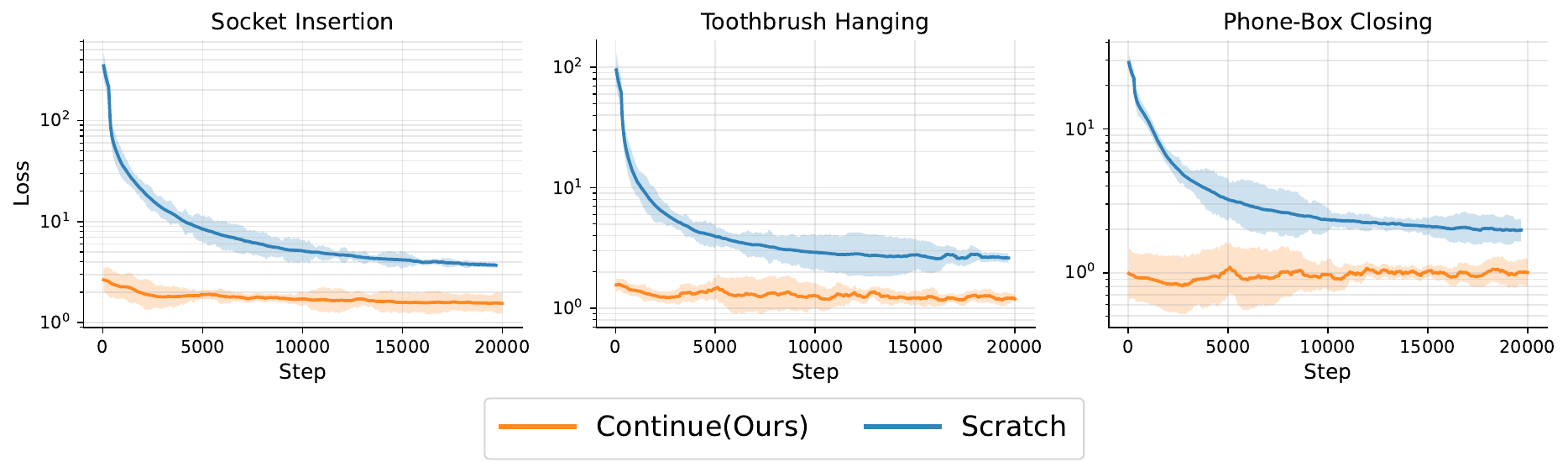}
        \vspace{-10pt}
        \caption{Log-scale distillation loss on the three previously learned tasks while learning the fourth task, with demo and replay buffers averaged for each task.}
        \label{fig:log loss scratch and continue} 
    \end{minipage}\hfill
    \begin{minipage}[t]{0.32\linewidth}
        \vspace{0pt}
        \vspace{-2pt}
        \centering
        {\tiny
        \setlength{\tabcolsep}{2.5pt}
        \renewcommand{\arraystretch}{1.35}
        \begin{tabular}{@{}lcc@{}}
            \multicolumn{3}{c}{\textbf{Success (\%)}} \\
            \hline
            Task & Cont. & Scratch \\
            \hline
            Socket Insertion & 100.0 & 94.0 $\pm$ 2.9  \\
            Toothbrush Hanging & 100.0 & 98.0 $\pm$ 1.2\\
            Phone-Box Closing & 100.0 & 79.0 $\pm$ 2.4\\
            Industrial Connectors Plugging & 100.0 & 100.0 \\
            \hline
        \end{tabular}}
        \vspace{15pt}
        \captionof{table}{Task success rates for the two initialization schemes, evaluated over three trial sets per task and 60 rollouts in total.}
        \label{tab:init_success_rates}
    \end{minipage}
    \vspace{-16pt}
\end{figure}

\paragraph{Adapter Gradient Routing}As shown in Figure~\ref{fig:adapter_routing}(a), our default routing blocks current-task actor gradients from updating the adapter while allowing historical distillation gradients to update it. In the fourth-task ablation, this asymmetric routing is critical: allowing both actor and distillation gradients to update the adapter prevents convergence by making the shared actor-critic representation nonstationary (Figure~\ref{fig:adapter_routing}(b,c)), while freezing the adapter for both streams yields increasing distillation loss, fewer than four retained tasks, and current-task learning time above 25 minutes (Figure~\ref{fig:adapter_routing}(d)). Continuing the default routing through the full six-task sequence completes all tasks with training time around 15 minutes (Figure~\ref{fig:adapter_routing}(e)).
\paragraph{Continual Initialization versus From-Scratch Learning}We conduct this initialization ablation during learning of the fourth task, after the policy has acquired three prior tasks, and compare two student initialization schemes under the same current-task training and historical distillation setup. In \emph{Continue}, the student is initialized from the previously consolidated policy, whereas in \emph{From scratch}, it is initialized with random parameters and must recover prior behaviors solely through historical distillation. As shown in Figure~\ref{fig:log loss scratch and continue} and Table~\ref{tab:init_success_rates}, continual initialization maintains consistently lower historical distillation loss on replay and demonstration buffers from the previously learned tasks while achieving higher task success rates. These results indicate that the consolidated policy provides a more stable basis for skill accumulation, and suggest that, as the number of tasks grows, one-shot merging after independently training each task may become increasingly ineffective.
\section{Limitations}
CIDER shows that, when single-task real-world RL can produce reliable policies, a robot can accumulate multiple manipulation skills by freezing the current consolidated policy as a teacher before learning each new task and distilling the teacher's behavior during subsequent online training. However, our current evaluation focuses on short-horizon manipulation tasks that can be solved within the available real-robot interaction budget. Longer-horizon tasks with sparse rewards, multi-stage contacts, or delayed failures may require additional exploration, reward design, or supervision before Interactive Distillation can be applied effectively.

\clearpage
\acknowledgments{If a paper is accepted, the final camera-ready version will (and probably should) include acknowledgments. All acknowledgments go at the end of the paper, including thanks to reviewers who gave useful comments, to colleagues who contributed to the ideas, and to funding agencies and corporate sponsors that provided financial support.}
\bibliography{example}  
\clearpage
\appendix

\section{Task and Implementation Details}
\paragraph{Task Descriptions.}
We evaluate our method on six real-robot manipulation tasks that require precise pose alignment, contact-rich interaction, and robust recovery from small initial variations.

\begin{figure}[htbp]
    \centering
    \begin{minipage}[t]{0.32\linewidth}
        \centering
        \includegraphics[width=\linewidth]{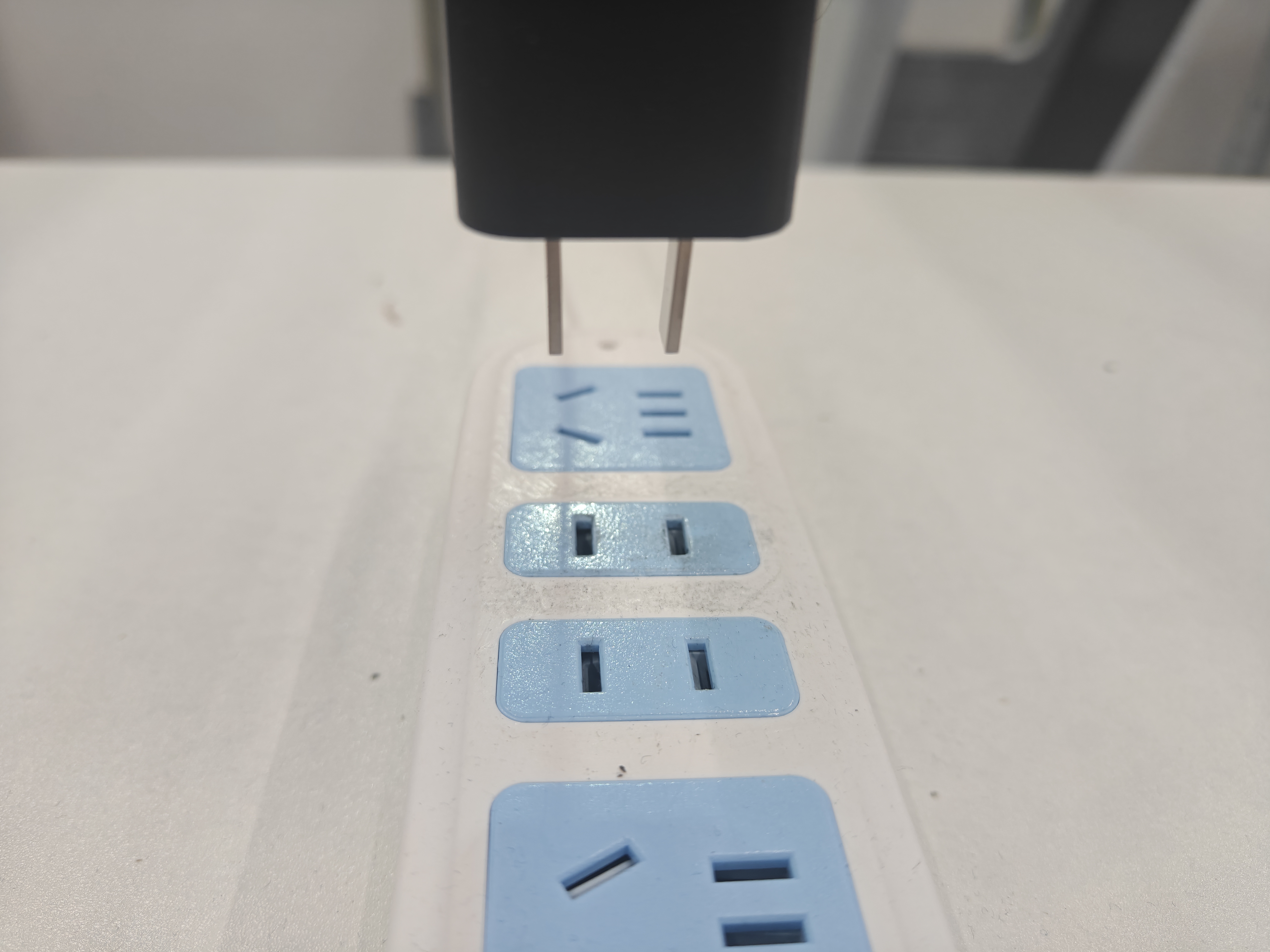}
        \caption*{(a) Socket Insertion}
    \end{minipage}
    \hfill
    \begin{minipage}[t]{0.33\linewidth}
        \centering
        \includegraphics[width=\linewidth]{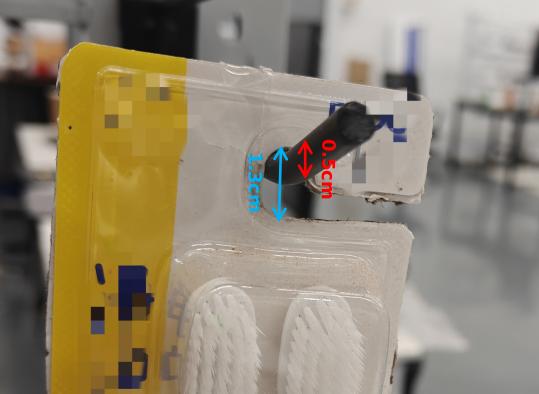}
        \caption*{(b) Toothbrush Hanging}
    \end{minipage}
    \hfill
    \begin{minipage}[t]{0.32\linewidth}
        \centering
        \includegraphics[width=\linewidth]{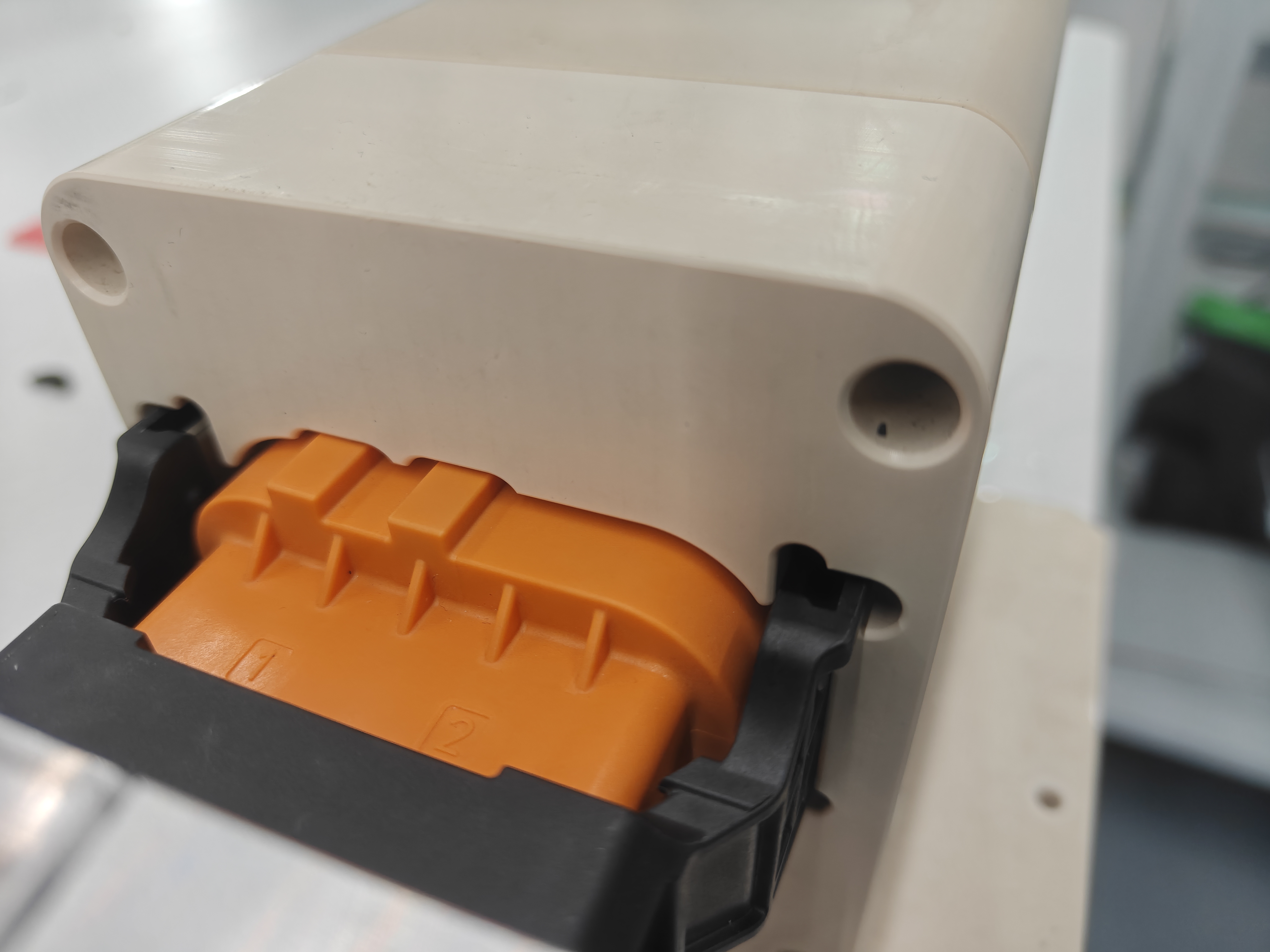}
        \caption*{(c) Industrial Connector Plugging}
    \end{minipage}
    \caption{Examples of tasks.}
    \label{task:toothbrush}
\end{figure}

\textbf{Socket Insertion.}
The robot picks up a plug and inserts it into a fixed socket. The episode starts with the plug placed on the table within a workspace region, while the socket remains fixed. Success is defined as the plug being fully inserted and mechanically seated in the socket. The main challenges are tight insertion tolerances, partial occlusion during contact, and the need to maintain a stable approach direction.

\textbf{Toothbrush Hanging.}
The robot grasps a packaged toothbrush and hangs it onto a narrow hook. As shown in Figure ~\ref{task:toothbrush}, the hanging aperture on the package is approximately $1.3$ cm in diameter, while the hook is only $0.5$ cm wide, leaving limited clearance for successful engagement. Success requires the package to remain stably suspended on the hook after release. This task is challenging because the robot must align the hanging hole with the hook while avoiding premature collision or slip.

\textbf{Industrial Connector Plugging.}
The robot inserts an industrial connector into its mating socket. The connector and socket are mounted in a constrained fixture, and the insertion direction must be accurately aligned before contact, with millimeter- to sub-millimeter-level tolerance. Success is defined as the connector being fully plugged into the socket. The task is difficult due to tight geometric constraints, contact-rich insertion dynamics, and limited visual feedback near the mating interface.

\textbf{Phone-Box Closing.}
The robot closes the lid of a smartphone packaging box. The phone packaging box measures $17.5 \times 8.6 $ cm, and its lid measures $17.8 \times 8.8$ cm, making lid closure sensitive to small pose errors due to the limited clearance. Success is achieved when the lid is fully closed without displacing the phone or the box. The task requires continuous pose correction and precise lid alignment during the closing motion.

\begin{figure}[htbp]
    \centering
    \begin{minipage}[t]{0.48\linewidth}
        \centering
        \includegraphics[width=\linewidth,height=4.5cm,keepaspectratio]{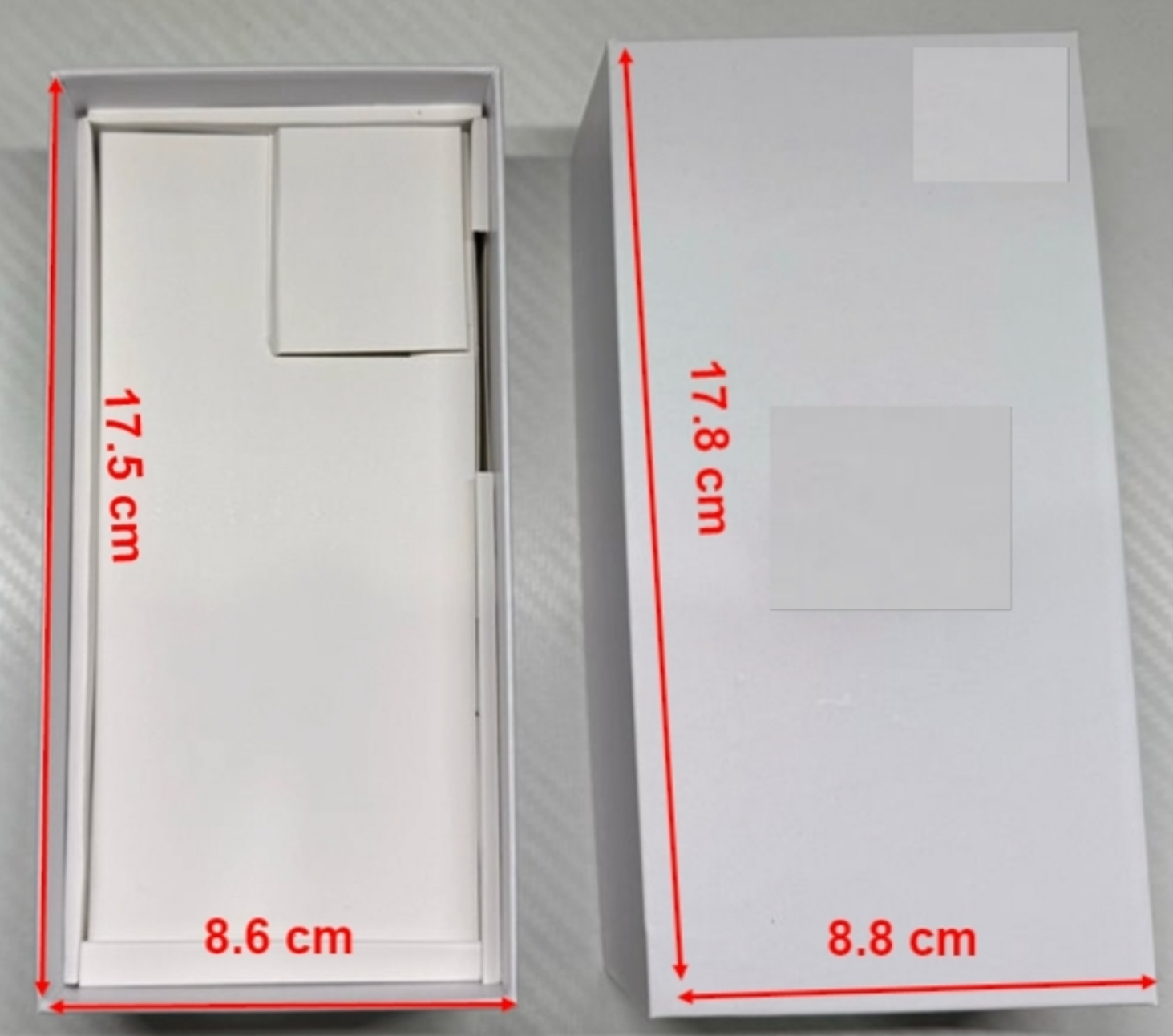}
        \caption*{(a) Phone-Box Closing}
    \end{minipage}
    \hfill
    \begin{minipage}[t]{0.48\linewidth}
        \centering
        \includegraphics[width=\linewidth,height=4.5cm,keepaspectratio]{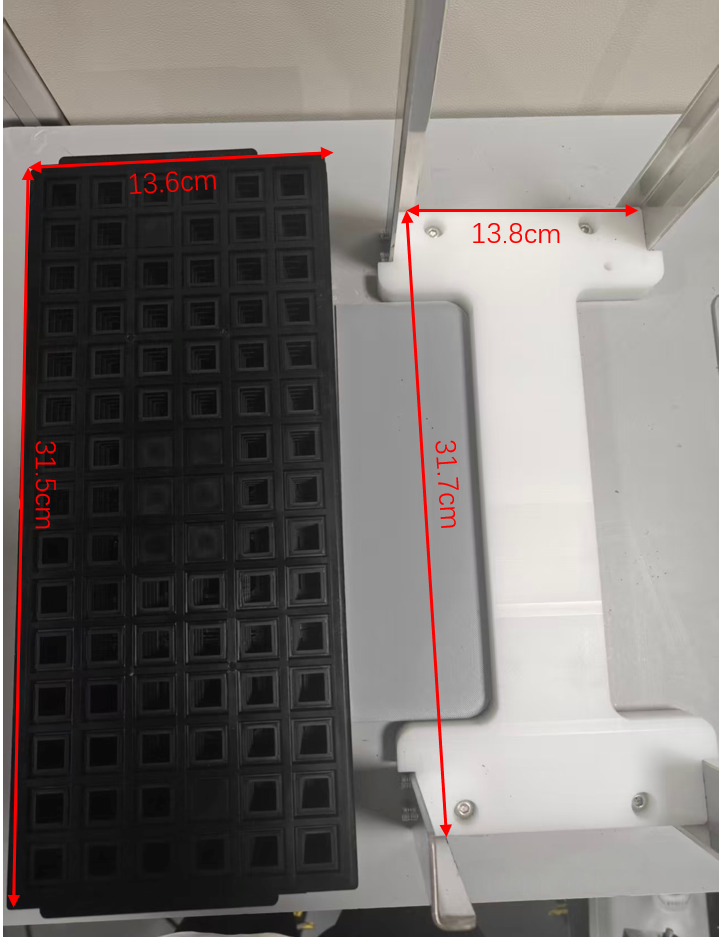}
        \caption*{(b) Chip-Tray Placement}
    \end{minipage}
    \caption{Examples of tasks.}
    \label{task:phone-box and chip-tray}
\end{figure}

\textbf{Chip-Tray Placement.}
The robot places a chip tray onto a designated test slot. The tray measures approximately $31.5 \times 13.6$ cm, while the slot measures $31.7 \times 13.8$ cm, leaving limited planar clearance for accurate placement. Success is defined as the tray being correctly seated on the test slot without tilting or disturbing the surrounding structure. The main challenges are precise planar alignment, gentle placement, and avoiding edge collisions during final insertion.

\begin{figure}[htbp]
    \centering
    \includegraphics[width=0.8\linewidth]{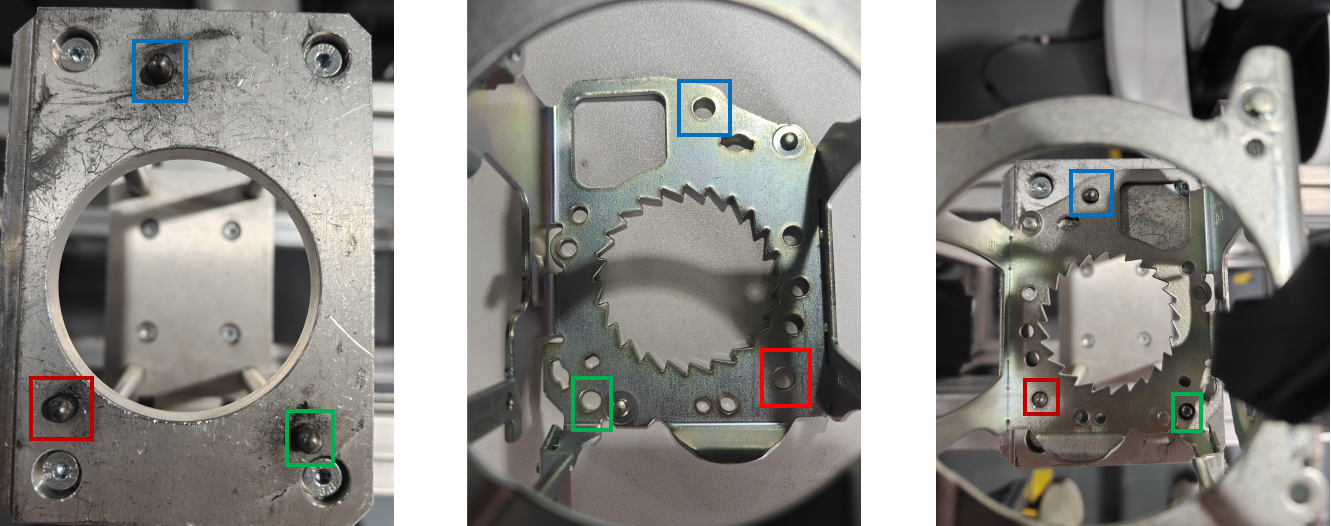}
    \caption{Safety Hook Coupling}
    \vspace{-8pt}
    \label{fig:Safety Hook Coupling}
\end{figure}

\textbf{Safety Hook Coupling.}
The robot engages a safety buckle with its corresponding receptacle through a three-pin coupling mechanism. As shown by the three colored markers in Figure \ref{fig:Safety Hook Coupling}, each pin must be aligned with its matching hole before insertion, resulting in sub-millimeter-level tolerance across the coupled interface. Success is achieved when all three pins are fully seated and the buckle remains locked after release. The task is challenging because it requires accurate approach orientation, controlled contact, and reliable engagement despite small pose errors.

The majority of the experimental parameters were identical for all experiments, including the following.
\begin{table}[htbp]
\centering
\caption{Common Experimental Parameters For All Tasks.}
\label{tab:training_parameters}
\begin{tabular}{l|l}
\hline
Parameter & Value \\
\hline
Action space & 6D twist \\
Reward function & Human-Assisted Binary classifier  \\
Initial offline demonstrations & 15 \\
Environment update frequency & 10 Hz \\
Maximum episode length & 200 environment steps \\
Reset method & Human reset \\
Proprioceptive encoder size & 64 \\
Critic MLP size & 256 $\times$ 256 \\
Number of critics & 2 \\
Motion policy MLP size & 256 $\times$ 256 \\
Discount factor & 0.96 \\
Automatic entropy tuning & True \\
Initial $\alpha$ & 0.01 \\
Target entropy & -3.5 \\
Optimizer & Adam \\
Learning rate & $3 \times 10^{-4}$ \\
Critic--actor ratio & 2 \\
Image augmentation & Random crop \\
\hline
\end{tabular}
\end{table}

The parameters that varied across experiments are summarized in Table ~\ref{varied}. Here, “mock” denotes using an all-black placeholder image to align the input format across all tasks.
\begin{table}[htbp]
\centering
\caption{Varied Experimental Parameters in Different Tasks.}
\label{tab:training_parameters}
\begin{tabular}{l|l|l}
\hline
Task & Observation space & Training steps  \\
\hline
Socket Insertion & wrist\_1, mock& 21000 \\
Toothbrush Hanging & wrist\_1, mock& 17000 \\
Phone-Box Closing & wrist\_1, mock& 28000 \\
Industrial Connectors Plugging & wrist\_1, mock& 29000 \\
Safety Hook Coupling & wrist\_1, mock& 25000 \\
Chip-Tray Placement & wrist\_1, wrist\_2& 23000 \\

\hline
\end{tabular}
\label{varied}
\end{table}

\section{Pseudocode}
\label{pseudocode}
\begin{algorithm}[H]
\caption{Interactive Distillation}
\label{alg:id}
\begin{algorithmic}[1]
\Require Scenario sequence $\mathcal{M}_{1:T}$; RLPD hyperparameters; update schedule $u_{\mathrm{RL}}$ and $u_{\mathrm{ID}}$ 
\Ensure Single deployable actor $\pi_{\theta_T}$
\State Initialize actor $\pi_\theta$, critics $Q_{\phi_1},Q_{\phi_2}$, and historical transition archives $\mathcal{B}_{<1}\leftarrow\emptyset$
\For{$k=1,\ldots,T$}
    \If{$k>1$}
        \State $\pi_{\mathrm{T}}^{k-1}\leftarrow\mathrm{stopgrad}(\pi_{\theta_{k-1}})$
        \State $\mathcal{C}_{<k}^{k-1}\leftarrow\{(s,\mu_{\mathrm{T}}(s),\sigma_{\mathrm{T}}(s)):(s,a,r,s')\in\bigcup_{i<k}\mathcal{B}_i\}$
        \State Initialize $\pi_\theta\leftarrow\pi_{\theta_{k-1}}$
    \EndIf
    \State Reset current replay buffer $\mathcal{D}_k$
    \While{the success criterion for $\mathcal{M}_k$ is not met}
        \State Roll out $\pi_\theta$ in $\mathcal{M}_k$ and append transitions to $\mathcal{D}_k$
        \For{$j=1,\ldots,u_{\mathrm{RL}}$}
            \State Update critics and actor using HIL-RL batches from $\mathcal{D}_k$
        \EndFor
        \If{$k>1$}
            \For{$j=1,\ldots,u_{\mathrm{ID}}$}
                \State Sample a task-balanced batch from $\mathcal{C}_{<k}^{k-1}$
                \State Update the actor using $\nabla_\theta\mathcal{L}_{\mathrm{distill}}^{k}$
            \EndFor
        \EndIf
    \EndWhile
    \State $\mathcal{B}_{k}\leftarrow\mathrm{Archive}(\mathcal{D}_k)$; set $\pi_{\theta_k}\leftarrow\pi_\theta$
\EndFor
\State \Return $\pi_{\theta_T}$
\end{algorithmic}
\end{algorithm}

%
%
%
%
%
%
%

\section{Policy Architecture and Deployment}
\subsection{Policy Architecture}
\label{app:policy_architecture}
All experiments, including CIDER and all baselines, use the same visuomotor policy architecture and deployment interface.

The policy takes two RGB images from different camera viewpoints as input. For tasks that require only a single viewpoint, the unused view is replaced with an all-zero placeholder image to maintain a consistent input dimensionality across tasks. Each image is processed by a ResNet-10 backbone pretrained on ImageNet, producing a feature map $X \in \mathbb{R}^{B \times C \times H \times W}$.

Each feature map is then passed through a lightweight visual adapter consisting of two sequential components. The first component is a Spatial Learned Embedding with $F{=}8$ learnable spatial masks $\{M_f\}_{f=1}^{F}$, where $M_f \in \mathbb{R}^{H \times W}$. For each sample $b$ and channel $c$, the embedding is computed as
\[
  e_{b,c,f} = \sum_{h,w} X_{b,c,h,w} M_{f,h,w},
\]
yielding a flattened spatial embedding of dimension $C \times F$. The second component projects this embedding to a 256-dimensional vector using a linear layer followed by LayerNorm and a Tanh activation. The two viewpoint-specific vectors are concatenated to form the joint visual representation $z \in \mathbb{R}^{B \times 512}$. This visual adapter is shared by the actor and critic heads. The gradient routing strategy for the visual adapter is determined by the training objective and is detailed in Section~\ref{ablation}.

The critic head concatenates $z$ with the action $a$ and feeds the result into a Q-network that outputs two independent Q-values for twin-critic estimation. The actor head maps $z$ through a two-layer MLP, where each layer consists of a linear projection, LayerNorm, and a Tanh activation with hidden dimension 256. The resulting feature is passed to two separate linear heads that predict the action mean $\mu_\theta(s)$ and log standard deviation $\log \sigma_\theta(s)$, which parameterize a diagonal Gaussian policy $\pi_\theta(\cdot \mid s)$.

\subsection{Deployment} 
The shared policy $\pi_\theta$ is conditioned only on visual observations and proprioceptive states, without any explicit task identifier fed into the network at test time. Task-relevant context is inferred implicitly from the visual scene, since each scenario presents a distinct object configuration and workspace layout that disambiguates the required manipulation behavior.

In multi-scenario deployment, task-specific preprocessing scripts configure camera crop regions and image normalization parameters for each scenario. We note that selecting the correct preprocessing configuration does require system-level knowledge of the current task: the operator or deployment system must specify which scenario is active in order to apply the appropriate camera parameters. Our setting is therefore not fully task-agnostic at the system level. A lightweight task identifier is present at deployment, but it operates exclusively as a perception pipeline selector and never enters the policy network as an input feature or conditioning signal.

This design reflects a practical real-world deployment constraint. In industrial and household settings, the physical arrangement of objects and fixtures typically makes the current task unambiguous to the operator, and configuring sensor parameters per scene is standard practice. Critically, the policy weights themselves receive no task label, and the same shared actor is deployed across all tasks without modification. Whether eliminating even this system-level task signal, for instance by learning to auto-detect the active scenario from raw observations, would preserve the same level of skill retention is an interesting direction for future work.

%
%
%
%
%
%
\section{Baseline Implementation Details}
\label{app:baseline_details}

All baselines share the same policy architecture (Appendix~\ref{app:policy_architecture}) and the HIL-RL interaction protocol as CIDER, differing only in the way data are used or how the optimization objective is defined. Unless noted otherwise, all RL hyperparameters are identical to those used for CIDER.

\textit{Finetune} is the simplest baseline for continual-learning. At the start of each new stage $k$, the current replay buffer is reset to empty, the policy is initialized from the parameters of the previous stage $\pi_{\theta_{k-1}}$, and HIL-RL runs exclusively on the current-task buffer $\mathcal{D}_k$ until the success criterion is met; historical transition data do not participate in training in any form. 

\textit{Finetune with Replay} extends this by retaining historical transition data and mixing them into the replay buffer: the merged buffer is used for both critic and actor updates, and the historical buffer is kept in full with no capacity limit.



\textit{Finetune with EWC} augments naive fine-tuning with an Elastic Weight Consolidation (EWC) regularization penalty~\cite{kirkpatrick2017overcoming}.
At the end of each task~$k$, the diagonal of the empirical Fisher Information Matrix (FIM) is estimated from the mini-batches(256) $N_{\mathrm{FIM}}=1{,}000$ sampled from the current-task buffer~$\mathcal{D}_k$:
\begin{equation}
    F_i^{(k)} = \frac{1}{N_{\mathrm{FIM}}}\sum_{n=1}^{N_{\mathrm{FIM}}} \left(\frac{\partial \log \pi_\theta(a_n \mid s_n)}{\partial \theta_i}\right)^2.
\end{equation}
The FIM diagonal and the consolidated parameter snapshot~$\theta^{*(k)}$ are stored and used as per-parameter importance weights in subsequent stages.
During training on task~$k' > k$, the total loss is:
\begin{equation}
    \mathcal{L} = \mathcal{L}_{\mathrm{BC}}(\theta;\,\mathcal{D}_{k'}) + \frac{\lambda}{|\theta|}\sum_{j=1}^{k'-1}\sum_i F_i^{(j)}\bigl(\theta_i - \theta_i^{*(j)}\bigr)^2,
\end{equation}
where the penalty coefficient $\lambda = 1.0$ and the sum is normalized by the total number of trainable parameters~$|\theta|$.
The EWC penalty is applied to all actor parameters.
Here, $\mathcal{D}_k$ consists of transitions during online RL training and does not include transitions from any previous tasks.

\textit{Sequential BC} and \textit{Sequential Offline RL} both update the policy offline without any online interaction with the real environment.

\textit{Sequential BC} initializes from $\pi_{\theta_{k-1}}$ and fine-tunes by minimizing the negative log-likelihood (NLL) of demonstrated transitions:
\begin{equation}
    \mathcal{L}_{\mathrm{BC}} = -\mathbb{E}_{(s,a)\sim\mathcal{D}_k}\bigl[\log\pi_\theta(a \mid s)\bigr],
\end{equation}
where $\mathcal{D}_k$ consists of demonstration transitions(human intervention) collected specifically for each task during the online training phase, and no historical demonstrations are included.
Training is run until convergence, for up to 10{,}000 gradient steps, using a batch size of 256, a learning rate of $3.0\times10^{-4}$, and the Adam optimizer.

\textit{Sequential Offline RL} likewise initializes from $\pi_{\theta_{k-1}}$ and applies SAC or TD3+BC to all transitions collected during the HIL-RL phase. Both variants are trained until the learning curves and evaluation metrics converge, for a maximum of 256{,}000 gradient steps. The SAC variant uses a batch size of 256 with automatic entropy tuning. The TD3+BC variant also uses a batch size of 256, with coefficient $\alpha = 10.0$ balancing the behavioral cloning term and the Q-value maximization objective. Target policy smoothing uses Gaussian noise with $\sigma=0.2$ clipped to $[-0.5, 0.5]$. Since TD3+BC achieved better real-world evaluation performance than SAC, we report the TD3+BC results in the Figure~\ref{fig:continual_results}.

Detailed hyperparameters are listed in Tables~\ref{tab:bc_ewc_params} and~\ref{tab:offline_rl_params}.

\begin{table}[htbp]
\centering
\caption{Hyperparameters for Finetune with EWC and Sequential BC.}
\label{tab:bc_ewc_params}
\begin{tabular}{l|c|c}
\hline
\textbf{Parameter} & \textbf{Finetune with EWC} & \textbf{Sequential BC} \\
\hline
BC loss & NLL & NLL \\
EWC $\lambda$ & 1.0 & --- \\
FIM batches $N_{\mathrm{FIM}}$ & 1{,}000 & --- \\
Penalty normalization & $1/|\theta|$ & --- \\
Training data & All transitions (demo buffer + replay buffer) & Demo buffer \\
Historical replay & None & None \\
\hline
\end{tabular}
\end{table}

\begin{table}[htbp]
\centering
\caption{Hyperparameters for Sequential Offline RL baselines.}
\label{tab:offline_rl_params}
\begin{tabular}{l|c|c}
\hline
\textbf{Parameter} & \textbf{SAC} & \textbf{TD3+BC} \\
\hline
Training steps/task & 256{,}000 & 256{,}000 \\
Auto entropy tuning & True & --- \\
TD3+BC $\alpha$ & --- & 10.0 \\
Policy noise $\sigma$ & --- & 0.2 \\
Noise clip & --- & 0.5 \\
Training data & All transitions & All transitions \\
Historical replay & None & None \\
\hline
\end{tabular}
\end{table}

\textit{DisCoRL} Six per-task expert policies are trained independently using the same HIL-RL pipeline as CIDER, each achieving a 100\% success rate over five evaluation runs with 20 trials each. After all experts are trained, we obtain a single student policy through offline distillation. The distillation stage follows the same procedure as CIDER, including task-balanced state sampling from each task's transitions and reverse-KL minimization between the student policy and the corresponding expert action distributions. Distillation is run for 125{,}000 training steps until policy convergence. Both the teacher and student policies use the same network architecture as CIDER.

\begin{table}[htbp]
\centering
\caption{Implementation Details for DisCoRL}
\label{tab:training_parameters}
\begin{tabular}{l|l}
\hline
Parameter & Value \\
\hline
Encoder & resnet10 \\
Latent Dimension & 64 \\
Teacher RL Algorithm & HIL-RL \\
\hline
Student Network & MLP(256,256) \\
Distillation Loss & reverse KL \\
Distillation Learning Rate & $3\times10^{-4}$ \\
Optimizer & Adam \\
Batch Size & 256 \\
Training Epochs & 150,000 \\
\hline
\end{tabular}
\end{table}

\paragraph{Main Results with STD}

Due to space constraints in the main paper, we report the mean success rates with standard deviations for Figure~\ref{fig:continual_results} here. Each result is evaluated over five independent evaluation sets, with 20 rollouts per set.
\begin{table}[H]
\centering
\caption{Success rates with STD across continual learning stages.}
\label{tab:continual_success_rates}
\resizebox{\textwidth}{!}{
\begin{tabular}{lcccccc}
\hline
\textbf{Method} & \textbf{Stage 1} & \textbf{Stage 2} & \textbf{Stage 3} & \textbf{Stage 4} & \textbf{Stage 5} & \textbf{Stage 6} \\
\hline
CIDER & $100.0 \pm 0.0$ & $100.0 \pm 0.0$ & $100.0 \pm 0.0$ & $100.0 \pm 0.0$ & $100.0 \pm 0.0$ & $100.0 \pm 0.0$ \\
DisCoRL & $100.0 \pm 0.0$ & $100.0 \pm 0.0$ & $100.0 \pm 0.0$ & $93.0 \pm 0.8$ & $55.8 \pm 1.5$ & $51.3 \pm 1.4$ \\
Finetune + Replay & $100.0 \pm 0.0$ & $81.5 \pm 2.5$ & $62.3 \pm 1.2$ & $56.5 \pm 1.8$ & $41.8 \pm 2.5$ & $26.2 \pm 2.9$ \\
Finetune + EWC & $100.0 \pm 0.0$ & $56.5 \pm 9.0$ & $38.7 \pm 2.4$ & $25.0 \pm 0.0$ & $20.0 \pm 0.0$ & $16.7 \pm 0.0$ \\
Finetune & $100.0 \pm 0.0$ & $50.0 \pm 0.0$ & $33.3 \pm 0.0$ & $25.0 \pm 0.0$ & $20.0 \pm 0.0$ & $16.7 \pm 0.0$ \\
Sequential Offline RL & $98.0 \pm 0.2$ & $42.0 \pm 0.6$ & $17.3 \pm 1.7$ & $25.0 \pm 0.0$ & $17.0 \pm 1.3$ & $16.2 \pm 1.2$ \\
Sequential BC & $90.0 \pm 0.6$ & $41.5 \pm 2.5$ & $24.7 \pm 0.7$ & $22.7 \pm 1.4$ & $13.6 \pm 1.7$ & $16.7 \pm 0.0$ \\
\hline
\end{tabular}
}
\end{table}

\paragraph{Analysis} CIDER builds on a line of policy-distillation methods for continual RL but 
differs structurally from each of them in ways that matter for real-world 
deployment.

Post-hoc consolidation (DisCoRL [4]). DisCoRL trains independent per-task 
experts and merges them into a single student via one offline distillation 
pass once all experts are available. This design has two limitations in our 
setting. First, it requires all experts to be simultaneously accessible, 
precluding incremental deployment as tasks arrive. Second, independently 
trained experts develop task-specific visual representations that are not 
aligned in parameter space. Consolidating them in a single offline pass subjects the student to conflicting gradient signals across tasks, a failure mode documented in the model-merging literature. This interference is compounded with the number of tasks: Figure~\ref{fig:continual_results} shows DisCoRL's stage-average success declining from 100\% to 51.3\% at stage 6,  with some early tasks collapsing (e.g., Phone-Box Closing drops to 25\%). CIDER avoids this by initializing each stage from the 
previously consolidated policy, so representations evolve incrementally and remain compatible across tasks. The teacher at each stage is already a 
multi-task policy rather than a collection of independently trained 
single-task experts.



\section{Additional Ablations}
\label{app:additional_ablations}


\paragraph{Distillation Update Ratio.}
We study the effect of the distillation update ratio $n=u_{\mathrm{RL}} : u_{\mathrm{ID}}$, where $n \in \{0.5, 1, 2\}$. All other training settings are held constant. Smaller values of $n$ increase the frequency of policy distillation and may improve retention of previously learned tasks, whereas larger values allocate a greater fraction of updates to current-task RL optimization. This experiment evaluates the trade-off between knowledge retention and adaptation to the current task as a function of distillation frequency. Due to time and cost constraints, we conduct ablations only on the two tasks (Phone-Box Closing and Industrial Connectors Plugging) that require the longest training time.
\begin{table}[htbp]
\centering
\caption{Distillation update-ratio ablation.}
\label{tab:ratio_ablation}
\begin{tabular}{c|c|c}
\hline
RL updates per distillation update & Final avg & Current-task speed \\
\hline
0.5 & 100 $\pm$ 0.0  & 34.86 min \\
1(Ours) & 100 $\pm$ 0.0 & 19.85 min \\
2 & 97.0 $\pm$ 2.0 & 19.88 min \\
\hline
\end{tabular}
\end{table}

\begin{figure}[htbp]
    \centering
    \includegraphics[width=0.6\linewidth]{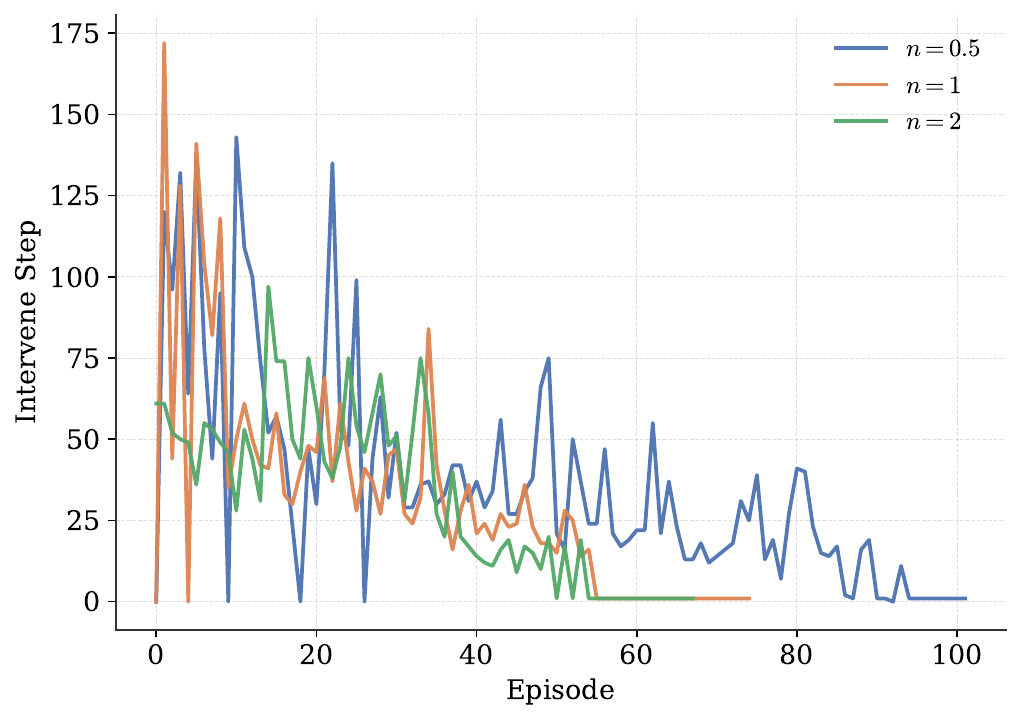}
    \caption{Intervened Steps}
    \vspace{-8pt}
    \label{fig:Intervened Steps}
\end{figure}

Table~\ref{tab:ratio_ablation} presents the effect of varying the distillation update ratio. Figure~\ref{fig:Intervened Steps} reports the number of human interventions across training episodes. We observe a clear trade-off between retention and adaptation speed to the current-task . Performing distillation after every RL update ($n=0.5$) yields the lowest forgetting but slows acquisition of the current task, suggesting that excessive rehearsal constrains policy adaptation. Conversely, larger ratios ($n=2$) result in slight forgetting. The intermediate setting ($n=1$) achieves the best overall balance, maintaining strong retention while preserving efficient learning on the current task. We therefore adopt $n=1$ as the default setting in all experiments.

\paragraph{Historical Task Sampling.}
For each distillation batch, previous-task transitions are sampled either uniformly from the union of all archived buffers, so that larger archives contribute proportionally more samples (dataset-proportional), or by first sampling a task uniformly at random and then sampling within that task's archive (task-balanced). Only the state component of each sampled transition is used for distillation. Task-balanced is our default; the contrast tests whether dataset-proportional sampling drowns out tasks with smaller archives. 
\begin{table}[htbp]
\centering
\caption{Historical sampling ablation.}
\label{tab:sampling_ablation}
\begin{tabular}{l|c}
\hline
Sampling & Final avg \\
\hline
Dataset-proportional & 78.0 $\pm$ 1.8 \\
Task-balanced (Ours) & 100 $\pm$ 0.0 \\
\hline
\end{tabular}
\end{table}

Results in Table~\ref{tab:sampling_ablation} show that task-balanced sampling consistently outperforms dataset-proportional sampling in terms of retention. While both strategies achieve comparable performance on the current task, dataset-proportional sampling tends to overrepresent tasks with larger replay archives, leading to reduced rehearsal of tasks with fewer stored transitions. In contrast, task-balanced sampling ensures that each previously learned task contributes equally to the distillation objective, resulting in improved historical task coverage and lower forgetting. Based on these observations, task-balanced sampling is used throughout the paper.

\paragraph{Task Order Robustness.}
We repeat the continual learning experiment using multiple random task permutations while keeping all training settings fixed. Results are reported in Table~\ref{tab:order_ablation}.

Different task orders lead to only minor performance variations. Although task ordering inevitably affects the sequence of policy updates, the overall retention and final performance remain stable across permutations, indicating that the proposed method is relatively insensitive to task order.

\begin{table}[htbp]
\centering
\caption{Task order robustness.}
\label{tab:order_ablation}
\begin{tabular}{l|c}
\hline
Sampling & Final avg \\
\hline
Ours$(M_1 \sim  M_6)$ & 100 $\pm$ 0.0 \\
$M_3, M_4, M_5, M_6, M_1, M_2$ & 96.2 $\pm$ 1.4 \\
$M_1, M_4, M_2, M_5, M_3, M_6$ & 98.4 $\pm$ 0.7 \\
\hline
\end{tabular}
\end{table}

\paragraph{Distillation Loss} Table~\ref{tab:loss_ablation} presents an ablation over different distillation objectives. We observe only minor differences among NLL, forward KL, and reverse KL, all of which achieve comparable performance on both current and previously learned tasks. In contrast, MSE action matching consistently underperforms in terms of retention. Due to time and cost constraints, we conduct ablations only on the first two tasks. 

One possible explanation is that NLL and KL-based objectives are naturally aligned with the probabilistic policy representation used by the actor. Rather than treating actions as deterministic regression targets, these objectives optimize quantities defined directly on the policy distribution, resulting in a training signal that is more consistent with the underlying policy parameterization. In contrast, MSE only penalizes differences between action values and ignores the probabilistic structure of the policy, which may lead to weaker preservation of previously learned behaviors.

Although reverse KL, forward KL, and NLL exhibit similar empirical performance, we adopt reverse KL as the default objective in subsequent experiments. Reverse KL explicitly encourages the student policy to place probability mass on actions favored by the teacher policy and therefore provides a principled objective for continual policy distillation. More importantly, the small performance differences among the probabilistic objectives indicate that the proposed framework is robust to the specific choice of distillation loss.

\begin{table}[htbp]
\centering
\caption{Toothbrush Hanging is evaluated as the current task after stage 2, and Socket Insertion is re-evaluated as the retention check.}
\label{tab:loss_ablation}
\begin{tabular}{l|c|c}
\hline
Loss & Toothbrush Hanging (current) & Socket Insertion (retention) \\
\hline
NLL & 100 $\pm$ 0.0 & 100 $\pm$ 0.0 \\
Forward KL & 100 $\pm$ 0.0 & 100 $\pm$ 0.0 \\
Reverse KL (Ours) & 100 $\pm$ 0.0 & 100 $\pm$ 0.0 \\
MSE & 100 $\pm$ 0.0 & 17.0 $\pm$ 1.2 \\
\hline
\end{tabular}
\end{table}

All evaluations are conducted over five runs, with 20 trials per run.

\section{Scalability Analysis}
\label{app:scalability_analysis}
CIDER scales linearly with $T$ in both buffer storage and cached teacher outputs $\mathcal{C}^{k-1}_{<k}$, while task-balanced sampling (Section~4.2) keeps per-update distillation cost constant, so the primary scaling bottleneck is buffer storage rather than per-step gradient computation. Whether a fixed-capacity shared actor can retain a substantially larger number of diverse skills remains an open question; our evaluation covers $T=6$ tasks, and whether representational capacity becomes a binding constraint for $T \gg 6$ is left to future work, with potential mitigations including incrementally increasing the adapter capacity or combining CIDER with structural expansion methods.

\end{document}